\documentclass{article}

\usepackage[preprint]{neurips_2025}

\usepackage[utf8]{inputenc} % allow utf-8 input
\usepackage[T1]{fontenc}    % use 8-bit T1 fonts
\usepackage{hyperref}       % hyperlinks
\usepackage{url}            % simple URL typesetting
\usepackage{booktabs}       % professional-quality tables
\usepackage{amsfonts}       % blackboard math symbols
\usepackage{nicefrac}       % compact symbols for 1/2, etc.
\usepackage{microtype}      % microtypography
\usepackage[table,dvipsnames]{xcolor}

\usepackage{multirow}
\usepackage{units}
\usepackage{pifont}
\usepackage{wrapfig}
\definecolor{darkgreen}{RGB}{0,120,0}
\definecolor{softgreen}{RGB}{0,100,0}
\usepackage{amsmath}
\usepackage{amssymb}
\usepackage{mathtools}
\usepackage{amsthm}
\usepackage{enumitem}
\usepackage{bm}

\newcommand{\eg}{\emph{e.g.}, }

\makeatletter
\def\thanks#1{\protected@xdef\@thanks{\@thanks
        \protect\footnotetext{#1}}}
\makeatother

\title{LongVie: Multimodal-Guided Controllable \\ Ultra-Long Video Generation}

\author{
\textbf{Jianxiong Gao$^{2,5}$, Zhaoxi Chen$^3$, Xian Liu$^4$} \\
\textbf{Jianfeng Feng$^2$, Chenyang Si$^{1,\dagger}$, Yanwei Fu$^{2,\dagger}$, Yu Qiao$^5$, Ziwei Liu$^{3,\dagger}$}
\thanks{$\dagger$: Co-corresponding authors.}\\
$^1$Nanjing University, $^2$Fudan University, $^3$S-Lab, Nanyang Technological University \\
$^4$NVIDIA, $^5$Shanghai AI Laboratory \\
\textcolor{orange}{\url{https://vchitect.github.io/LongVie-project/}}
}
\begin{document}

\maketitle

\begin{figure}[h]
\centering
\vskip -0.25in
\includegraphics[width=\textwidth]{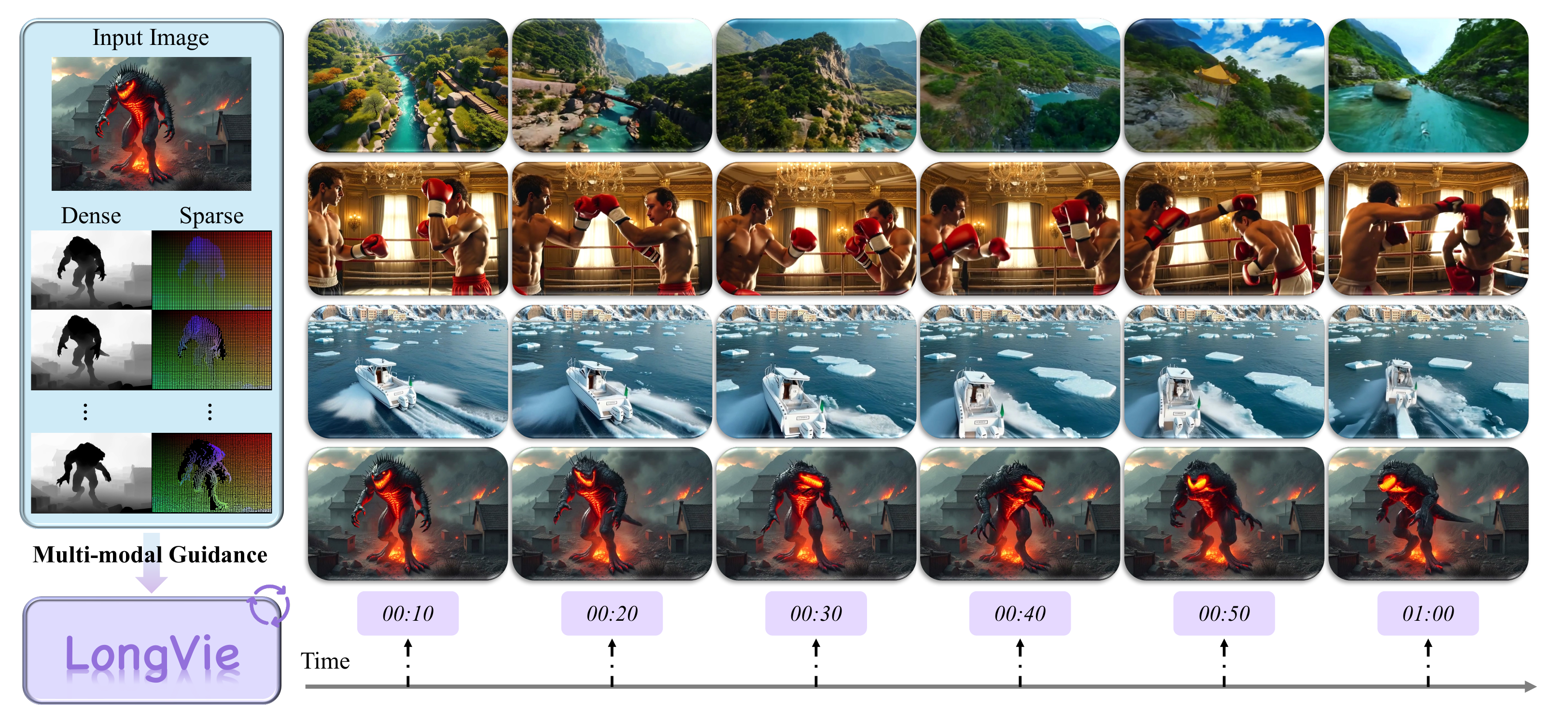}
\caption{\textbf{LongVie} is a controllable ultra-long video generation framework guided by both dense and sparse control signals, with a degradation-aware training strategy to balance the contribution of the modalities. It applies global normalization to the control signals and employs unified initialization noise to autoregressively generate videos lasting up to one minute.}
\label{fig:teaser}
\vskip -0.05in
\end{figure}

\begin{abstract}

Controllable ultra-long video generation is a fundamental yet challenging task. Although existing methods are effective for short clips, they struggle to scale due to issues such as temporal inconsistency and visual degradation.
In this paper, we initially investigate and identify three key factors: separate noise initialization, independent control signal normalization, and the limitations of single-modality guidance. To address these issues, we propose \textbf{LongVie}, an end-to-end autoregressive framework for controllable long video generation. LongVie introduces two core designs to ensure temporal consistency: \textbf{1)} a unified noise initialization strategy that maintains consistent generation across clips, and \textbf{2)} global control signal normalization that enforces alignment in the control space throughout the entire video. To mitigate visual degradation, LongVie employs \textbf{3)} a multi-modal control framework that integrates both dense (\eg depth maps) and sparse (\eg keypoints) control signals, complemented by \textbf{4)} a degradation-aware training strategy that adaptively balances modality contributions over time to preserve visual quality. We also introduce \textbf{LongVGenBench}, a comprehensive benchmark consisting of 100 high-resolution videos spanning diverse real-world and synthetic environments, each lasting over one minute. Extensive experiments show that LongVie achieves state-of-the-art performance in long-range controllability, consistency, and quality.

\end{abstract}

\section{Introduction}

Recent advancements in video generation have been significantly driven by the availability of large-scale datasets and the development of powerful generative architectures, particularly diffusion models~\cite{ho2020DDPM, rombach2021highresolution, opensora, meta_moviegen, dataverse}. These innovations have greatly enhanced the video generation, enabling models such as CogVideoX~\cite{yang2024cogvideox}, HunyuanVideo~\cite{kong2024hunyuanvideo}, Kling~\cite{kling}, Sora~\cite{sora}, and Wanx2.1~\cite{wan2025} to generate high-quality videos with text prompts.

Despite these successes, a major challenge in video generation remains the ability to achieve precise control, ensuring that the generated video aligns seamlessly with the user’s creative vision. Recent efforts have attempted to overcome this limitation by integrating control framework~\cite{zhang2023controlvideo, chen2023control, guo2024sparsectrl, ma2024follow, huang2025fine, gu2025das} into the generation process in order to achieve more refined and customizable outputs. However, these methods primarily focus on the controllable generation of short video clips, with limited exploration into the challenges associated with generating longer, controllable videos, such as those lasting up to one minute. In practical applications, the ability to generate long, coherent, and visually consistent videos is critical, yet it remains a fundamentally complex and unresolved problem.

\begin{figure}
\begin{center}
\includegraphics[width=0.98\columnwidth]{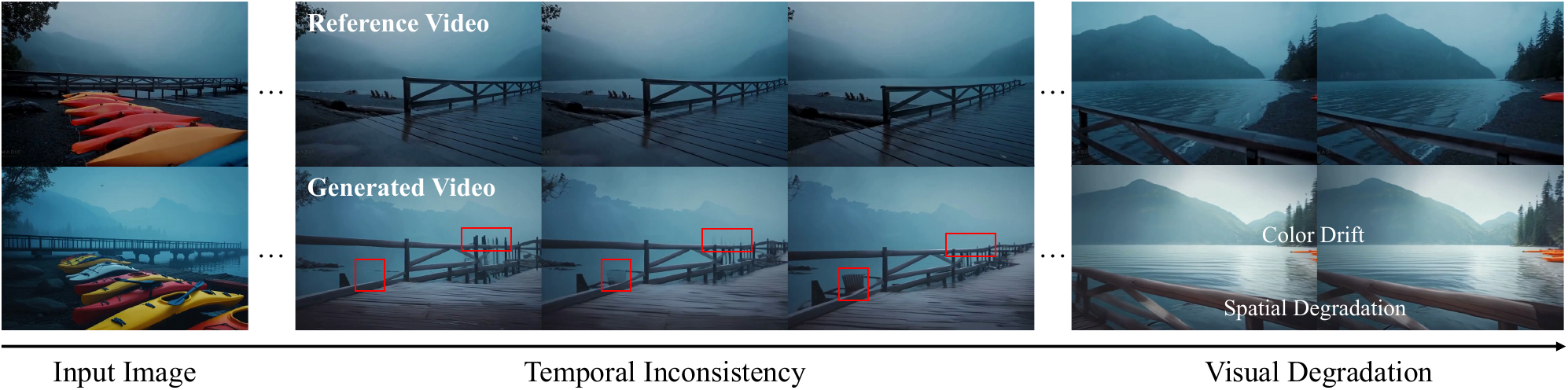}
\vskip -0.05in
\caption{\textbf{Temporal Inconsistency and Quality Degradation.} These are the primary limitations encountered when applying current controllable models to long video generation.}
\label{fig:intro_problem}
\end{center}
\vskip -0.15in
\end{figure}

In this paper, we focus on the problem of controllable long video generation. Directly generating one-minute-long controllable videos typically requires considerable computational resources. Hence, we begin by analyzing the applicability of autoregressive generation to long video synthesis, wherein short video segments are generated sequentially, each initialized from the final frame of the preceding segment. While this approach offers a more tractable solution, it exhibits notable limitations when extended to continuous long-form generation. As shown in Fig.~\ref{fig:intro_problem}, as the video length increases with each successive iteration, two main aspects become evident: \textbf{1) temporal inconsistency} between adjacent clips, resulting in noticeable scene transitions or flickering quality; \textbf{2) visual degradation}, where each subsequent clip exhibits reduced overall visual fidelity compared to the previous one. These two aspects highlight the challenges in maintaining both visual continuity and high quality over extended video sequences when relying solely on autoregressive mechanism.

Building upon this analysis, we further investigate the underlying causes of these two issues. We find that temporal inconsistency primarily stems from two factors: the independent normalization of control signals across clips and the separate noise initialization employed in each generation step. Per-clip normalization disrupts global coherence in the control space, leading to inconsistencies between adjacent frames—especially when the control signals (\eg depth or motion maps) lack sufficient scene-level alignment. Additionally, varying noise initialization inputs across clips introduces unpredictable variations in content, further weakening temporal continuity. On the other hand, visual degradation is largely attributed to the limitations of single-modality control. Dense signals provide fine-grained structural guidance but may dominate generation process, while sparse signals offer high-level semantics but lack adequate spatial detail. Over time, reliance on a single modality causes accumulated errors and a progressive decline in visual output quality.

To address both challenges, we propose \textbf{LongVie}, a controllable long video generation framework built on an autoregressive paradigm. LongVie introduces two core components to enhance temporal consistency: \textbf{1) unified noise initialization}, which ensures consistent generative dynamics across clips, and \textbf{2) global control signal normalization}, which maintains alignment in the control space throughout the entire video. These two techniques work together to reduce abrupt transitions and flickering artifacts, resulting in a temporally coherent generation process. To mitigate visual degradation, LongVie employs a \textbf{3) multi-modal control framework} that integrates both dense (\eg depth maps) and sparse (\eg keypoints) modalities for complementary guidance, along with a \textbf{4) degradation-aware training scheme} that balances their contributions over time. Through these designs, LongVie enables the generation of minute-long videos that are not only temporally consistent but also maintain high visual fidelity under fine-grained controllability.

To evaluate the effectiveness of controllable long video generation, we construct a dataset of 100 high-quality videos, each with a duration of at least one minute. The videos cover a wide range of scenes, including both real-world environments and game-based scenarios, ensuring diversity in content and motion. We refer to this dataset as \textbf{LongVGenBench}. We evaluate LongVie on LongVGenBench in terms of long-term visual quality and temporal consistency. The results show that LongVie achieves leading performance in controllable long video generation.

Our main contributions are as follows:

\begin{itemize} 
\item We present a comprehensive analysis of the limitations in existing controllable video generation models for long videos, identifying two key challenges: long-term temporal inconsistency and visual degradation. To address these issues, we propose \textbf{LongVie}, the first autoregressive framework for controllable long video generation.
\item To improve visual quality, we propose a multi-modal control mechanism that integrates dense and sparse control signals to exploit their respective advantages, alongside a degradation-aware training strategy that balances their contributions.
\item To enhance temporal consistency, we leverage unified noise initialization and global control signals normalization for a world-consistent generative dynamics over time steps.
\item We introduce \textbf{LongVGenBench}, an evaluation dataset for controllable long video generation, comprising 100 diverse, high-quality videos, each lasting at least one minute.
\end{itemize}

\section{Methodology}

\textbf{Preliminary: Diffusion Models.}
Diffusion models are a strong type of video generator that turn random noise into realistic videos by learning how to remove the noise step by step. So, the model $\epsilon_\theta$ learns to undo a process that adds noise to data. During training, it is trained to predict the noise added at each step:
\begin{align}
    \label{eqn_loss}
    \mathcal{L} = \mathbb{E}_{\boldsymbol{x},\epsilon \sim \mathcal{N}(0,1),t} \Big [  \lVert \epsilon - \epsilon_\theta(\boldsymbol{x}_{t}, t) \rVert^2_2 \Big ],
\end{align}
where $\boldsymbol{x}_t$, $\epsilon$, $\mathbf{x}_0$ denote the noisy data at time $t$,  the true noise, and the original data, respectively.
To save computing power, Latent Diffusion Models~\cite{rombach2021highresolution} work in a smaller, compressed space. An autoencoder turns $\boldsymbol{x}$ into a latent code $\boldsymbol{z} = \mathcal{E}(\boldsymbol{x})$, and the model learns to predict noise in $\boldsymbol{z}_t$.

\textbf{Overview.}
We extend CogVideoX~\cite{yang2024cogvideox} with a ControlNet-style architecture~\cite{controlnet} to incorporate the external control signals. A lightweight control branch, partially shared with the base model, processes the control signals. While effective for short video synthesis, most controllable diffusion-based models, including CogVideoX and its variants, are not designed to handle long-duration generation, such as one-minute sequences. Generating such long videos in a single forward pass is computationally prohibitive. As a result, a common practice is to generate videos in an autoregressive manner—producing short clips sequentially, with each initialized from the final frame of the previous one. In our implementation, we follow this approach using a depth-conditioned variant of CogVideoX~\cite{yang2024cogvideox}. However, as discussed in the following subsection and illustrated in Fig.~\ref{fig:intro_problem}, this strategy introduces two major challenges: (1) temporal inconsistency across consecutive clips, and (2) progressive quality degradation due to accumulated errors over time.

\subsection{Rethinking Controllable Generation of Long Videos}
\label{sec2.2}

\begin{wrapfigure}{r}{0.55\textwidth}
  \centering
  \vskip -0.15in
  \includegraphics[width=0.55\textwidth]{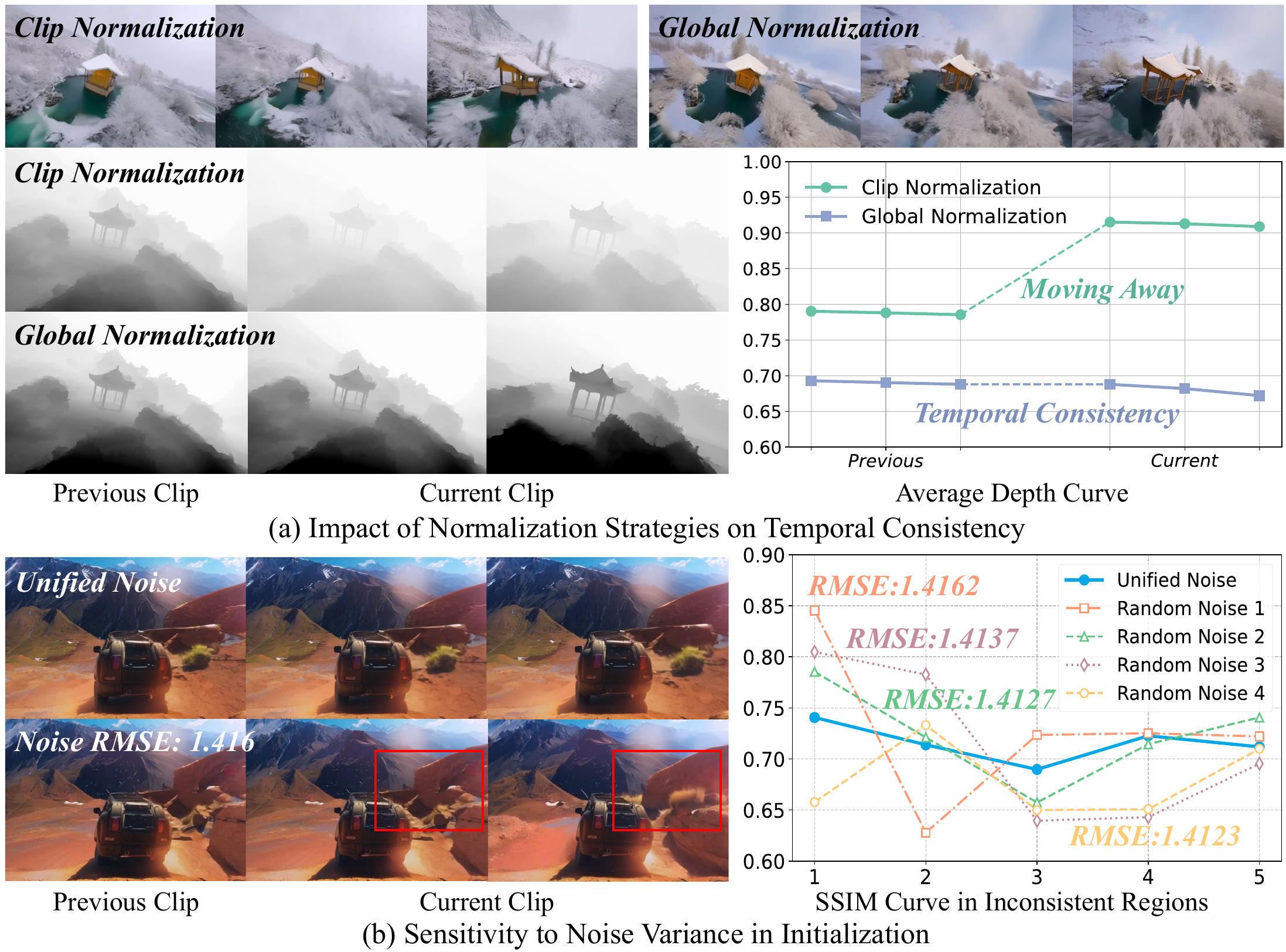}
  \vskip -0.1in
\caption{\textbf{Analysis of temporal inconsistency.} Temporal inconsistency arises from clip-wise normalization and random noise initialization, while global normalization and unified noise alleviate these issues.}
  \label{fig:temporal_analyse}
  \vskip -0.05in
\end{wrapfigure}

\textbf{Temporal Inconsistencies.}
To investigate the source of temporal inconsistencies, we analyze the input signals used in controllable video generation models. In models that rely on external control, such as depth maps (see Fig.~\ref{fig:temporal_analyse} (a)), these signals are usually normalized independently for each clip. We find that this per-clip normalization introduces inconsistencies across clips. For example, the same scene may appear with different depth values in consecutive clips. As a result, the model receives mismatched guidance across clips, which distorts its perception of scene geometry and motion continuity. This leads to temporal artifacts such as unnatural zooming or abrupt viewpoint changes (Fig.\ref{fig:temporal_analyse} (a)). 
These findings suggest that independent normalization breaks the alignment of control signals across clips, especially when the signals lack global context or a consistent reference scale, ultimately causing visible inconsistencies between clips.

Inspired by our analysis of control signal normalization, we further examine the impact of noise initialization on temporal consistency. In diffusion-based video generation, the initial noise plays a critical role in determining the overall structure and motion of the output. We observe that temporal inconsistencies frequently occur at the beginning of each generated clip, suggesting a strong correlation between noise initialization and temporal disruptions. In standard autoregressive generation, each clip is sampled from a different random noise input. This variation introduces inconsistencies in motion, appearance, or scene layout across clips, even when the control signals remain aligned. 
Our empirical study, illustrated in Fig.~\ref{fig:temporal_analyse} (b), confirms this effect: Clips with larger differences in initialization noise—measured by the root mean squared error (RMSE) relative to that of the first clip—tend to exhibit more noticeable temporal inconsistencies, as shown by the Structural Similarity Index (SSIM) curves in Fig.~\ref{fig:temporal_analyse} (b).

In summary, temporal inconsistencies in long video generation are primarily caused by misaligned control signals across clips and variations in initialization noise, both of which disrupt the continuity of motion and appearance.

\begin{figure}[h]
\begin{center}
\vskip -0.05in
\includegraphics[width=\columnwidth]{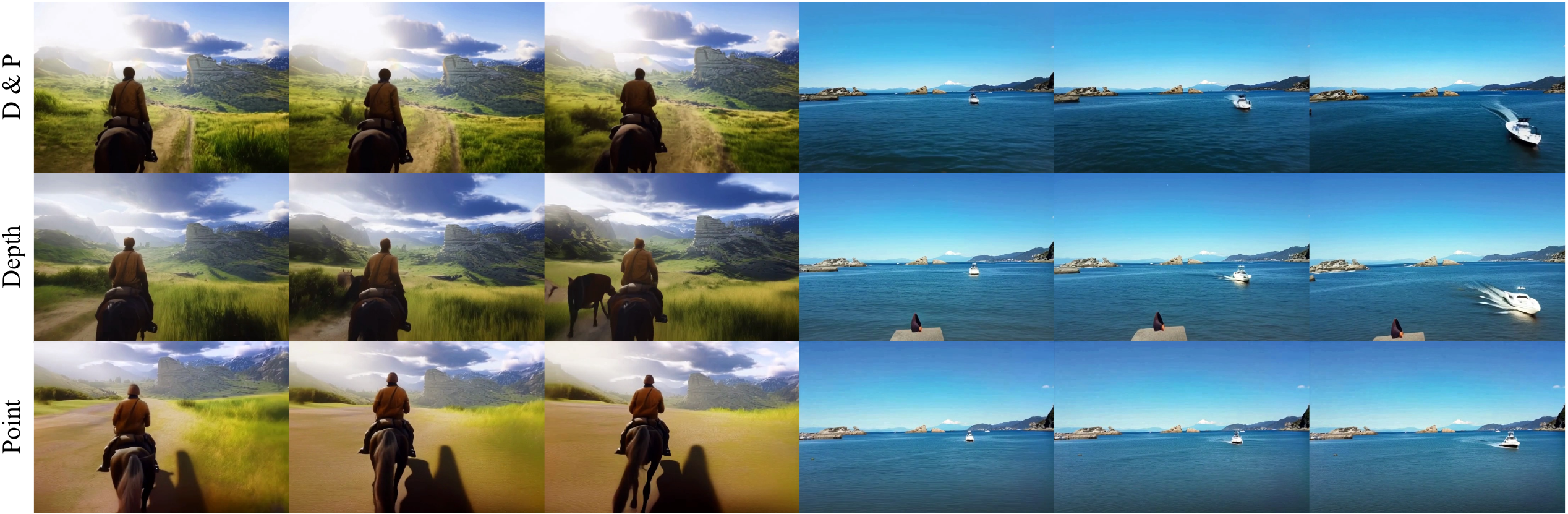}
\vskip -0.05in
\caption{\textbf{Visual degradation caused by single-modality control.} Depth-only and point-only settings show noticeable degradation, while D\&P, combining both modalities, alleviates the issue.}
\label{fig:modal_analyze}
\end{center}
\vskip -0.1in
\end{figure}

\textbf{Visual Quality Degradation.}
Building on our analysis of temporal inconsistency, we further investigate the issue of visual quality degradation in long video generation. Controlling long video generation using per-frame signals is a practical strategy to maintain stability and visual quality. However, different control modalities come with inherent trade-offs that limit their effectiveness over extended sequences. Taking depth as an example of a dense modality, it provides pixel-level structural information across frames. While effective for preserving local geometry, it offers limited control over nearby or distant regions and lacks the capacity to represent high-level semantics such as object motion or scene dynamics. As shown in Fig.~\ref{fig:modal_analyze}, these limitations result in artifacts and degraded quality, especially in complex scenes.
In contrast, point-based control is a sparse modality that captures semantic cues by specifying a few keypoints. While it effectively guides motion and object structure, its sparse nature makes it sensitive to scene changes and less reliable in maintaining semantic alignment across frames. These limitations show that neither dense nor sparse control alone is sufficient for consistent long video generation. When control signals fail to align with the evolving scene, visual quality degrades progressively.

\begin{figure}
\begin{center}
\includegraphics[width=\columnwidth]{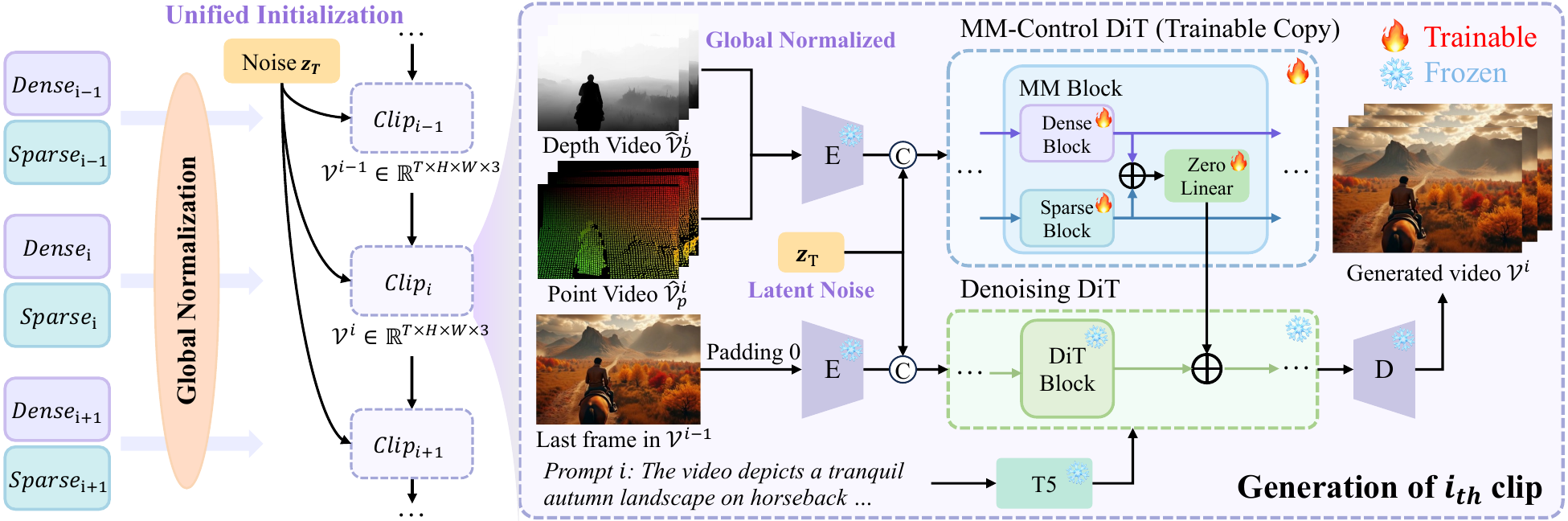}
% \vskip -0.05in
\caption{\textbf{Framework of LongVie.} We adopt dense and sparse control signals as scene guidance to generate controllable long videos in an autoregressive manner. We also apply global normalization and unified initialization noise to improve temporal consistency during the generation process.}
\label{fig:framework}
\end{center}
\vskip -0.15in
\end{figure}

\subsection{LongVie Framework}
\label{sec2.3}

To address the aforementioned challenges, we propose \textbf{LongVie}, a framework for controllable long video generation, as illustrated in Fig.~\ref{fig:framework}.
At the core of our model is a multi-modal framework that leverages both dense and sparse controls to guide the scene effectively. 
In the following paragraphs, we will first introduce the general framework, and then present the effective training and inference strategies to solve the issues mentioned in Sec.~\ref{sec2.2}.

\textbf{Multi-Modal Control Injection.}
Specifically, we adopt depth maps as dense control signals and point maps as sparse signals, leveraging the detailed structural information provided by depth and the high-level semantic cues captured by point trajectories. To construct the point map sequences, we follow the procedure in DAS~\cite{gu2025das}, where a set of keypoints is tracked across frames and colorized according to their depth values. 

\begin{figure}
\begin{center}
\includegraphics[width=\columnwidth]{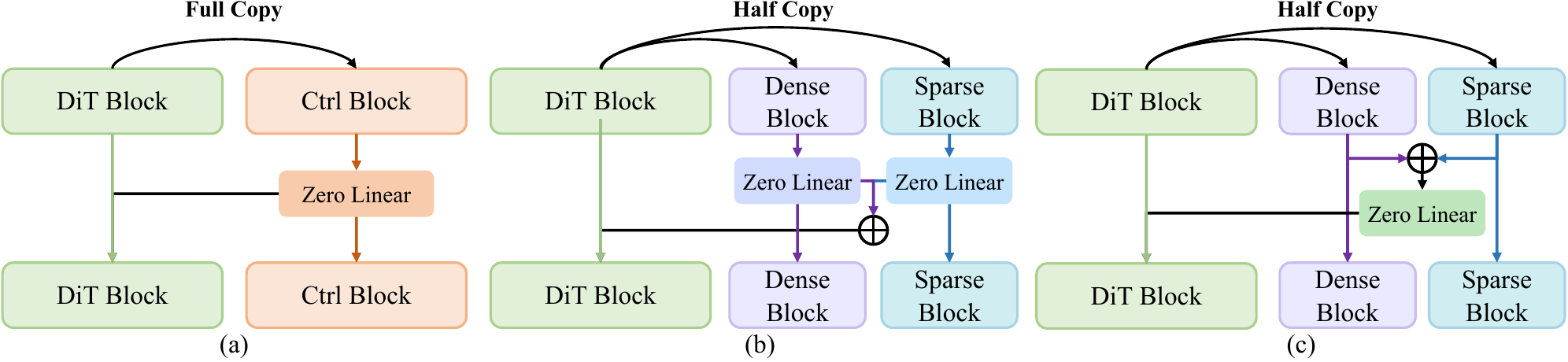}
% \vskip -0.12in
\caption{\textbf{Variants of Multi-Modal Control Integration.} 
Compared to the standard ControlNet design (a), we present two variant structures (b) and (c) that integrate dense and sparse control signals.}
\label{fig:control_block_vis}
\end{center}
\vskip -0.2in
\end{figure}

Inspired by the ControlNet~\cite{controlnet} architecture, we build our Multi-Modal Control DiT by duplicating the initial layers of the pre-trained CogVideoX DiT and incorporating multi-modal conditioning inputs while keeping the base model frozen. Specifically, we freeze the parameters $\theta$ of the original DiT blocks and construct two trainable branches, each corresponding to a distinct control modality: dense (depth) and sparse (point). These branches, denoted as $\mathcal{F}_\text{D}(\cdot;\theta_\text{D})$ and $\mathcal{F}_\text{P}(\cdot;\theta_\text{S})$ , are lightweight copies of the original DiT layers with parameters $\theta_\text{D}$ and $\theta_\text{P}$ , respectively. Each processes the corresponding encoded control signal $\boldsymbol{c}_\text{D}$ or $\boldsymbol{c}_\text{P}$. To integrate the control branches into the base generation path, we adopt zero-initialized linear layers $\phi^l$, as shown in Fig.~\ref{fig:control_block_vis}, similar in spirit to the zero convolutions used in ControlNet~\cite{controlnet}. We compare two variant structures and adopt (b) in this paper for its empirical stability.
These layers allow the control signals to be injected additively into the main generation stream without affecting the model’s initial behavior, as they output zero at the start of training. The overall computation in the $l$-th controlled DiT block is defined as:
\begin{align}
    \label{eqn_fuse}
    \boldsymbol{z}^l = \mathcal{F}^l( \boldsymbol{z}^{l-1}) + \phi^l(\mathcal{F}_\text{D}^l(\boldsymbol{c}^{l-1}_\text{D}) + \mathcal{F}_\text{P}^l(\boldsymbol{c}^{l-1}_\text{P}) ),
\end{align}
where $\mathcal{F}^l$ is the frozen base DiT block, and $\mathcal{F}_\text{D}^l$, $\mathcal{F}_\text{P}^l$ are the control-specific sub-blocks.

In practice, we duplicate the first 18 blocks of the original CogVideoX DiT to construct the Multi-Modal Control DiT. The remaining blocks remain untouched to preserve the generative capacity of the base model. During training, only the parameters in the control branches and the fusion layers $\phi^l$ are updated, while the backbone remains fixed. This setup allows us to introduce strong, flexible conditioning from both dense and sparse modalities, while maintaining the stability and pre-trained knowledge of the original model.

\textbf{Global Normalization.}
To reduce temporal inconsistencies caused by independently normalized control inputs, we adopt a global normalization strategy for the depth video. Specifically, we compute the 5th and 95th percentiles of all pixel values across the entire video sequence and use these as the global minimum and maximum normalization bounds. The depth values are then clipped to this range and linearly scaled to [0, 1]. This percentile-based normalization is robust to outliers and ensures that the depth values across all clips are on a consistent scale. As shown in Fig.~\ref{fig:temporal_analyse} (top-right), this global normalization effectively reduces inter-clip variations and leads to more temporally aligned control signals. After normalization, the depth video is segmented into overlapping clips to match the autoregressive inference process and facilitate corresponding point map extraction.

\textbf{Unified Noise Initialization.}
To further enhance temporal consistency, we use a shared noise initialization across all video segments during generation. Instead of sampling a different noise vector for each clip, we sample a single noise instance and apply it uniformly across the entire sequence. This unified noise serves as a consistent latent prior, reducing variations between adjacent clips that typically arise from independently sampled noise. As illustrated in Fig.~\ref{fig:temporal_analyse} (bottom right), this approach significantly improves temporal coherence, mitigating flickering and promoting smooth transitions throughout the generated video.

\textbf{Degradation Strategy for Modal Balance.}
While multi-modal control offers the potential for richer and more accurate video generation, simply combining dense and sparse control signals does not guarantee improved performance. In practice, we observe that dense signals such as depth tend to dominate the generation process, often suppressing the semantic and motion-level guidance provided by sparse signals like keypoints. This imbalance can lead to suboptimal visual quality, particularly in scenarios requiring high-level semantic alignment over time.

To address this issue, we propose a degradation-based training strategy designed to regulate the relative influence of dense control signals and encourage more balanced utilization of both modalities. This strategy weakens the dominance of the dense input through controlled perturbations at both the feature and data levels:

\textit{1) Feature-level degradation:}  
During training, with probability $\alpha$, we randomly scale the latent representation of the dense control input by a factor sampled uniformly from the range $[0.05, 1]$. Accordingly, Equation~\ref{eqn_fuse} can be reformulated as:
\begin{align}
    \label{eqn_fuse}
    \boldsymbol{z}^l = \mathcal{F}^l( \boldsymbol{z}^{l-1}) + \phi^l(\alpha \cdot \mathcal{F}_\text{D}^l(\boldsymbol{c}^{l-1}_\text{D}) + \mathcal{F}_\text{P}^l(\boldsymbol{c}^{l-1}_\text{P}) ),
\end{align}
This operation reduces the magnitude of the dense features, making the model more reliant on complementary information provided by the sparse modality. Over time, this encourages the network to learn a more balanced integration of both control sources.

\textit{2) Data-level degradation:}  
Given a dense control tensor $D \in \mathbb{R}^{B \times C \times H \times W}$, we apply degradation with probability $\beta$ using two techniques:
\textit{a) Random Scale Fusion:}  
A set of spatial scales $\{1, 1/2, \dots, 1/2^n\}$ is predefined. One scale is randomly excluded, and the remaining scales are used to generate downsampled versions of the input, which are then upsampled to the original resolution. Each version is assigned a random weight normalized to sum to 1. Their weighted sum forms a fused depth map with multi-scale degradation and randomness, enhancing robustness to spatial variation.
\textit{b) Adaptive Blur Augmentation:}  
An average blur with a randomly chosen odd-sized kernel is applied to the dense input to reduce sharpness, limiting the model’s tendency to overfit to local depth details.

Together, these degradations prevent over-reliance on dense signals and improve the model’s ability to integrate complementary information from sparse modalities, ultimately enhancing long-term video quality and consistency.

\subsection{Versatility to Downstream Video Generative Tasks}
\label{sec2.4}
In this section, we describe how LongVie can be adapted to various long video generation tasks.

\begin{figure}
\begin{center}
\includegraphics[width=\columnwidth]{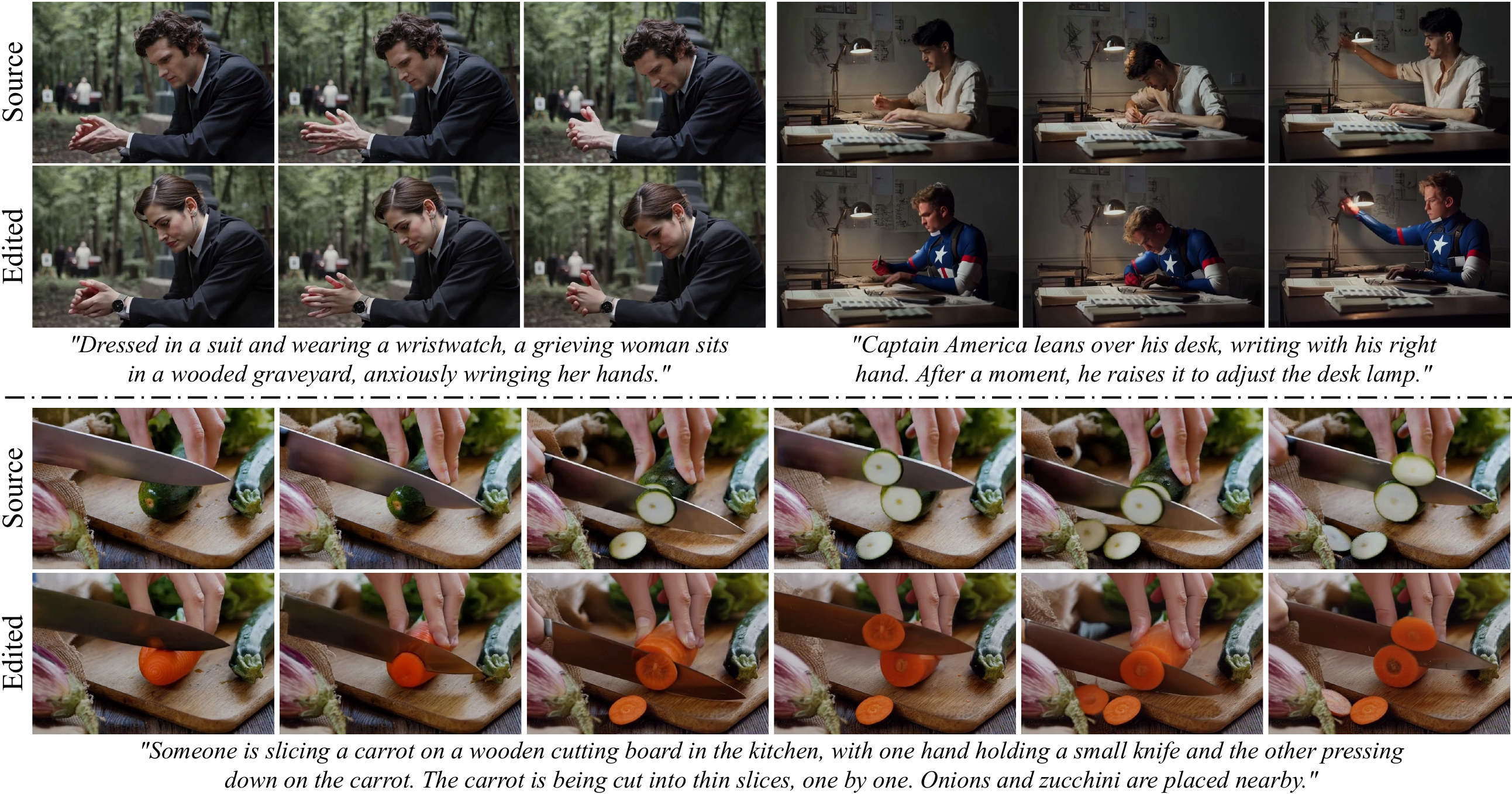}
\vskip -0.05in
\caption{\textbf{Video Editing.} LongVie enables high-quality and temporally consistent video editing.}
\label{fig:video_inpainting}
\end{center}
\vskip -0.2in
\end{figure}

\begin{figure}
\begin{center}
\includegraphics[width=\columnwidth]{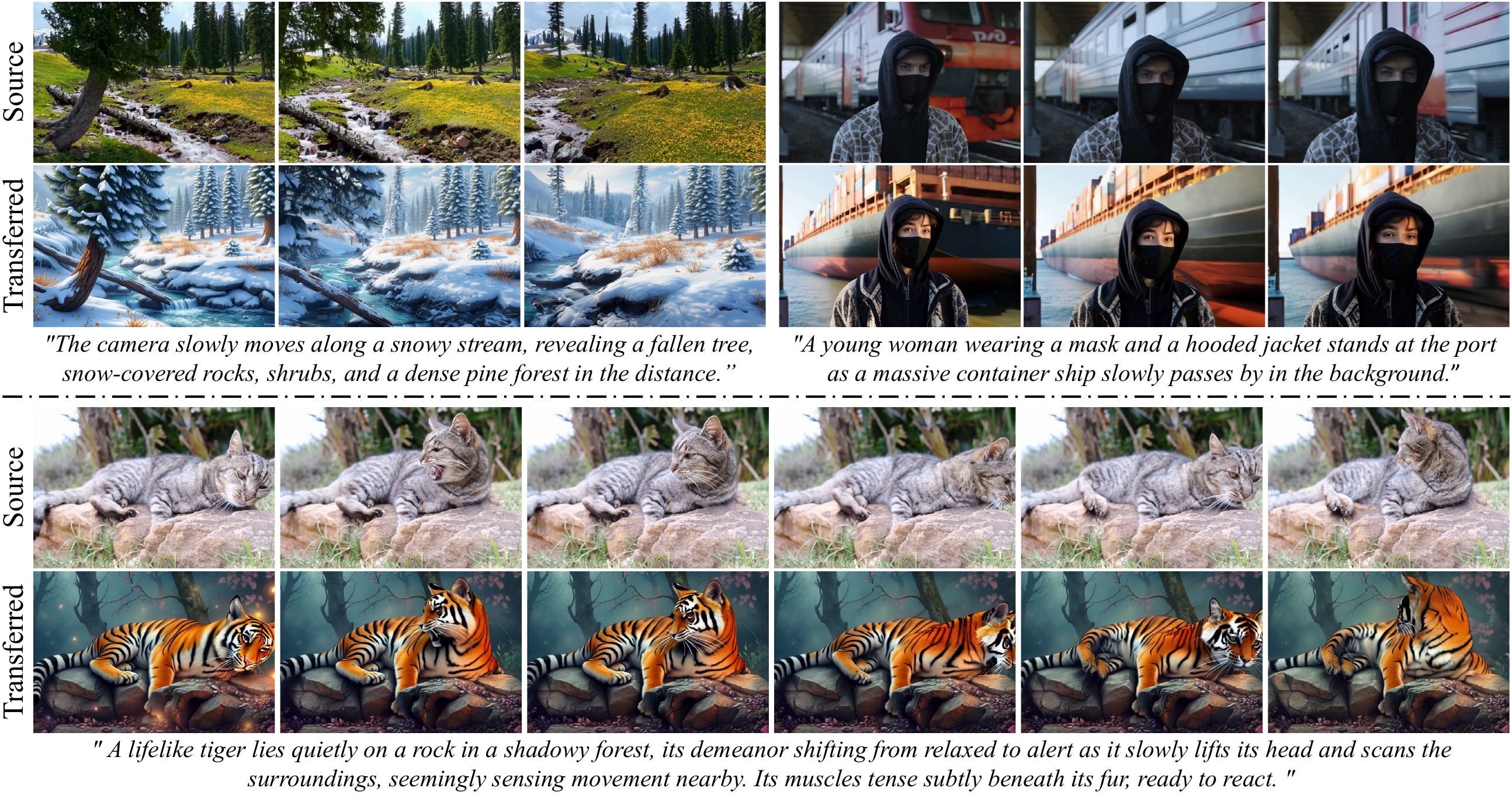}
\vskip -0.05in
\caption{\textbf{Motion \& Scene Transfer.} LongVie effectively transfers motion and scene across frames.}
\label{fig:motion_transfer}
\end{center}
\vskip -0.25in
\end{figure}

\begin{figure}
\begin{center}
\includegraphics[width=\columnwidth]{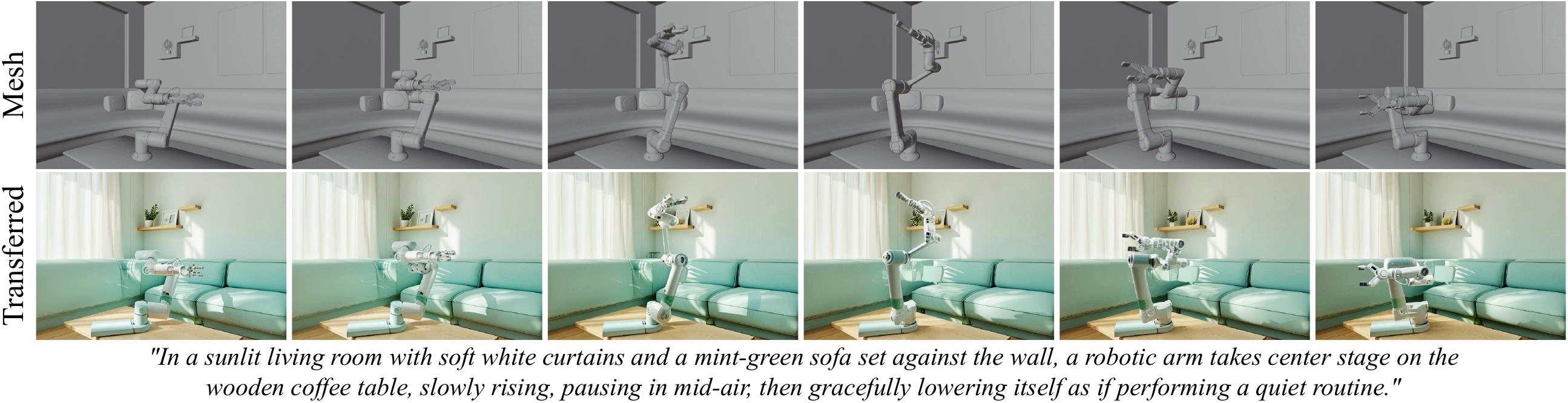}
\vskip -0.1in
\caption{\textbf{Mesh-to-Video.} LongVie generates realistic long videos from animated 3D meshes.}
\label{fig:mesh2video}
\end{center}
% \vskip -0.2in
\end{figure}

\textbf{Video Editing.} 
LongVie can be adapted for long-range video editing. We first edit the initial frame by selecting a target region and completing it using the fill model from FLUX~\cite{flux2024}. The completed frames are then fed into LongVie, along with dense and sparse control signals, to generate temporally consistent edited videos.

\textbf{Motion \& Scene Transfer.}  
LongVie supports motion and scene transfer across long videos. Given a source video with target motion or layout, the depth-to-image model from FLUX~\cite{flux2024} is used to synthesize an initial frame reflecting the desired attributes. Depth and point maps are extracted as control signals for LongVie, which generates videos that preserve the transferred motion or scene while ensuring temporal and visual consistency.

\textbf{Controllable Mesh-to-Video.}
LongVie generates long videos from animated 3D meshes without textures. We render the mesh in a 3D engine (\eg Blender) to produce an animation. A depth-to-image model~\cite{flux2024} synthesizes an initial stylized frame, and depth maps and point trajectories are extracted from the animation. These signals guide LongVie to produce coherent, high-quality videos, enabling seamless integration of animated 3D assets into photorealistic domains.

\section{Experiments}
\noindent \textbf{Implementation Details.}
\label{experiments}
We implement LongVie by copying and fine-tuning 18 DiT blocks in each model. During training, we first extract depth maps using Video Depth Anything~\cite{video_depth_anything} as the dense control signal, and then apply SpatialTracker~\cite{SpatialTracker} to track 3D points based on the normalized depth. Following DAS~\cite{gu2025das}, we uniformly sample 4,900 points per short video as sparse control signals. Each training video is divided into 49-frame clips at a resolution of 480×720 and 8 frames per second (fps).
We then use Qwen2.5-VL-7B~\cite{Qwen2.5-VL} to automatically generate captions for the training videos. In total, we use 130,000 videos to train LongVie. 
The training data consists of ACID~\cite{acid}, Vchitect\-T2V\-DataVerse~\cite{dataverse}, and MovieNet~\cite{huang2020movienet}, with detailed information provided in the supplementary material.
LongVie is trained using the AdamW optimizer with a learning rate of 1e-4. Training is conducted on 8 A100 GPUs over approximately 3,000 iterations, with an effective batch size of 64, and takes about 5 days to complete.
During inference, LongVie requires approximately 4.5 minutes to sample a 6-second video. Consequently, generating a one-minute controllable video takes about 45 minutes on a single A100 GPU.

\subsection{Qualitative and Quantitative Results}

\noindent \textbf{LongVGenBench.}
To address the lack of suitable benchmarks for controllable long video generation, we introduce LongVGenBench, a dataset of 100 one-shot videos, each lasting at least one minute at 1080p resolution. Existing datasets are inadequate, as they lack long, continuous, one-shot videos—crucial for evaluating temporal consistency and controllability.
LongVGenBench spans diverse real-world and game-based scenarios, and includes challenging cases such as rapid scene transitions and complex motions (see supplementary material for details), making it a strong benchmark for this task.
For evaluation, each video is divided into 6-second clips, and captions are automatically generated using Qwen-2.5-VL-7B~\cite{Qwen2.5-VL} to serve as prompts. Each video is further segmented into ten 49-frame clips at 8 frames per second (fps), with a 1-frame overlap, following the autoregressive setup used in our experiments. Control signals are extracted from the split clips.
During validation, no transformation is applied to the first frame of each video, ensuring fair comparison and enabling accurate assessment of generation quality, as ground-truth frames are available for reference.

\begin{table} 
\small
\centering
\setlength{\tabcolsep}{1.2mm}{
\caption{
\textbf{Quantitative results of LongVie and baselines on our LongVGenBench.} DAS-LV and Depth-LV refer to the adapted versions of DAS and depth-controlled CogVideo, respectively, for long video generation. Go-With-Flow refers to the model Go-With-The-Flow. 
\textbf{Bold} indicates the best performance, and \underline{underline} denotes the second-best.
\label{tab:metrics}
}
\vskip -0.05in
{\begin{tabular}{lccccccccc}
\toprule
\textsc{Methods} & S.C.$\uparrow$ & B.C.$\uparrow$ & O.C.$\uparrow$ & D.D$\uparrow$ & T.F.$\uparrow$ & A.Q.$\uparrow$ & I.Q.$\uparrow$ & SSIM$\uparrow$ & LPIPS$\downarrow$ \\
\midrule
\midrule
\rowcolor{gray!10}
CogVideoX & 85.38\% & 90.46\% & 20.72\% & 22.06\% & 97.80\% & \underline{54.96\%} & \underline{64.89\%} & 0.374 & 0.521 \\
\rowcolor{gray!10}
StreamingT2V & 83.18\% & 90.56\% & 20.58\% & 21.15\% & 97.52\% & 52.51\% & 62.45\% & 0.360 & 0.572 \\
VideoComposer~\cite{wang2023videocomposer} & 80.33\% & 88.83\% & 19.83\% & 27.78\% & 96.36\% & 52.83\% & 59.33\% & 0.346 & 0.583 \\
Motion-I2V~\cite{shi2024motion} & 84.25\% & 89.32\% & 19.99\% & 37.34\% & 97.16\% & 53.26\% & 61.57\% & 0.385 & 0.504 \\
Go-With-Flow~\cite{burgert2025gowiththeflow} & 84.37\% & 90.62\% & \underline{21.79\%} & \underline{46.15\%} & 97.77\% & 53.59\% & 62.21\% & 0.453 & 0.394 \\
DAS~\cite{gu2025das} & 86.06\% & 90.78\% & 21.10\% & 36.76\% & \underline{98.11\%} & 53.28\% & 64.57\% & 0.401 & 0.482 \\
Depth-LV & \underline{87.09\%} & \underline{91.37\%} & 21.25\% & 46.06\% & 97.70\% & 54.80\% & 64.84\% & \underline{0.508} & \underline{0.347} \\
\rowcolor{cyan!15}
\textbf{LongVie} & \textbf{87.12\%} & \textbf{91.76\%} & \textbf{21.82\%} & \textbf{46.59\%} & \textbf{98.43\%} & \textbf{55.31\%} & \textbf{64.91\%} & \textbf{0.557} & \textbf{0.290} \\
\midrule
\bottomrule
\end{tabular}}}
\vskip -0.05in
\end{table}

\textbf{Evaluation Metrics and Baselines.}
To evaluate the effectiveness of LongVie, we adapt several video generation models for long video generation, including the base model CogVideoX~\cite{yang2024cogvideox}; the controllable models VideoComposer~\cite{wang2023videocomposer}, Motion-I2V~\cite{shi2024motion}, Go-With-The-Flow~\cite{burgert2025gowiththeflow}, and DAS~\cite{gu2025das}; as well as a depth-controlled variant of CogVideoX, termed Depth-LV.  
We also compare against StreamingT2V~\cite{StreamingT2V}, a strong image-driven baseline for long video generation.
For evaluation, we follow the widely used benchmark VBench~\cite{vbench} and employ seven metrics—\textit{Background Consistency}, \textit{Subject Consistency}, 
\textit{Overall Consistency}, \textit{Temporal Style}, \textit{Dynamic Degree}, \textit{Temporal Flickering}, and \textit{Imaging Quality}—to assess temporal coherence and visual fidelity. We also report traditional similarity-based metrics, including SSIM and LPIPS, to quantify the reconstruction quality of generated videos with respect to their input references.

\textbf{Experimental Results.}
The quantitative results in Tab.~\ref{tab:metrics} demonstrate that LongVie achieves the best temporal consistency and controllability among all baselines, achieving state-of-the-art performance.
To further illustrate the effectiveness of LongVie in controllable long video generation, we present video editing results in Fig.~\ref{fig:video_inpainting}, where LongVie faithfully replaces target characters or objects as specified. The results of motion and scene transfer are shown in Fig.~\ref{fig:motion_transfer}, indicating that LongVie can handle complex motions and scene transformations. Additionally, we showcase controllable mesh-to-video generation results in Fig.~\ref{fig:mesh2video}. We first place the desired animated 3D models in Blender and repaint them using FLUX. As shown, LongVie successfully synthesizes high-quality videos from the repainted meshes.

\textbf{User Study.}
To comprehensively evaluate the models, we carefully design and conduct a subjective user study. To mitigate participant fatigue, we structure the evaluation accordingly. From the generated videos, we randomly select 80 samples, each paired with its corresponding prompt and control signals. The evaluation focuses on five key aspects: Visual Quality, Prompt-Video Consistency, Condition Consistency, Color Consistency, and Temporal Consistency. We compare five models: CogVideoX~\cite{yang2024cogvideox}, StreamingT2V~\cite{StreamingT2V}, DAS-LV~\cite{gu2025das}, Depth-LV, and LongVie. A total of 60 participants are invited. For each evaluation aspect, participants rank the outputs of the five models, assigning 5 points to the best and 1 point to the worst. The average scores across all evaluations are summarized in Tab.~\ref{tab:userstudy}. As shown, our proposed method, LongVie, achieves the highest scores across all criteria.

\begin{table}
\centering
\small
% \vskip -0.05in
\setlength{\tabcolsep}{3.5mm}{
\caption{\textbf{User Study Comparison with Baselines.} We define 5 dimensions and invite 60 participants to evaluate LongVie with the baselines. LongVie outperforms all baselines in every aspect.}
\label{tab:userstudy}
\vskip -0.05in
\begin{tabular}{lccccc}
\toprule
\multirow{2}{*}{\textsc{Methods}} & Visual          & Prompt-Video      & Condition   &  Color     &  Temporal  \\
                                  & Quality         & Consistency       & Consistency & Consistency & Consistency\\
\midrule
\midrule
CogVideoX~\cite{yang2024cogvideox}      & 2.232 & 2.251 & 1.967 & 2.514 & 2.272 \\
StreamingT2V~\cite{StreamingT2V}   & 2.054 & 2.017 & 2.232 & 1.942 & 1.959 \\
DAS-LV~\cite{gu2025das}         & 3.072 & 3.138 & 3.253 & 3.035 & 3.183 \\
Depth-LV       & 3.286 & 3.153 & 3.318 & 3.267 & 3.262 \\
\rowcolor{cyan!15}
\textbf{LongVie} & \textbf{4.387} & \textbf{4.471} & \textbf{4.282} & \textbf{4.298} & \textbf{4.365} \\
\midrule
\bottomrule
\end{tabular}}
\vskip -0.15in
\end{table}

\subsection{Ablation Study}

\textbf{Unified Initial Noise \& Global Normalization.}
We observe that both the unified initialization of noise and the normalization of control signals significantly affect the consistency and quality of the generated videos. To evaluate their impact, we generate videos under three settings: without Global Normalization, without Unified Initial Noise, and without both. The results, reported in Tab.~\ref{tab:ablation_noise_norm} and evaluated using four corresponding metrics, demonstrate that both Global Normalization and Unified Initial Noise contribute positively to controllable long video generation.

\textbf{Degradation Training Strategy.}
We conduct an ablation study on the degradation-aware training strategy to balance the contributions of multiple modalities. Results in Tab.~\ref{tab:ablation_noise_norm} show that both feature and data level strategies improve the visual quality of long video generation.

\section{Related Works}

\textbf{Video Diffusion Models}.
Recently, video diffusion models have advanced rapidly, enabling the generation of high-resolution videos with consistent appearance and motion across dozens of frames. Representative open-source models include CogVideoX~\cite{yang2024cogvideox}, HunyuanVideo~\cite{kong2024hunyuanvideo}, and Wan2.1~\cite{wan2025}. In parallel, several closed-source models such as Sora~\cite{sora} and Kling~\cite{kling} have also demonstrated impressive results.
While these models can generate high-quality videos from text or image prompts, they still exhibit limitations in fine-grained controllability and scalability to long-duration content. To bridge this gap, our work focuses on controllable long video generation. Specifically, we build upon CogVideoX~\cite{yang2024cogvideox} as our base model. Importantly, our proposed LongVie framework can be seamlessly extended to enable controllable long video generation using any existing video diffusion architecture.

\begin{table} 
\small
\centering
\setlength{\tabcolsep}{2.3mm}{
\caption{\textbf{Ablation study for our proposed components.} The \textcolor{cyan!15}{\rule{1em}{1.5ex}} block denotes experiments targeting temporal consistency, while the \textcolor{purple!15}{\rule{1em}{1.5ex}} block denotes those focusing on visual quality.}
\label{tab:ablation_noise_norm}
\vskip -0.08in
{\begin{tabular}{lcccc}
\toprule
\multirow{2}{*}{\textsc{Methods}} & \multicolumn{2}{c}{Consistency}    &  \multicolumn{2}{c}{Quality} \\
                                  & Subject         & Background       & Aesthetic & Imaging \\
\midrule
\midrule
Full model  &  \cellcolor{cyan!15} \textbf{87.12\%} & \cellcolor{cyan!15} \textbf{91.76\%} & \cellcolor{purple!15}\textbf{54.91\%} & \cellcolor{purple!15}\textbf{64.91\%} \\
\midrule
~ \textit{w/o} Global Normalization  & 
\cellcolor{cyan!15}86.39\%~(\textcolor{softgreen}{$\downarrow$0.73}) & 
\cellcolor{cyan!15}91.41\%~(\textcolor{softgreen}{$\downarrow$0.35}) & 
54.88\%~(\textcolor{softgreen}{$\downarrow$0.03}) & 
64.81\%~(\textcolor{softgreen}{$\downarrow$0.10}) \\	

~ \textit{w/o} Unified Initial Noise &  
\cellcolor{cyan!15}86.63\%~(\textcolor{softgreen}{$\downarrow$0.49}) & 
\cellcolor{cyan!15}91.59\%~(\textcolor{softgreen}{$\downarrow$0.17}) & 
54.80\%~(\textcolor{softgreen}{$\downarrow$0.11}) & 
64.84\%~(\textcolor{softgreen}{$\downarrow$0.07}) \\

~ \textit{w/o} Both                 & 
\cellcolor{cyan!15}86.23\%~(\textcolor{softgreen}{$\downarrow$0.89}) & 
\cellcolor{cyan!15}91.37\%~(\textcolor{softgreen}{$\downarrow$0.39}) & 
54.78\%~(\textcolor{softgreen}{$\downarrow$0.13}) & 
64.86\%~(\textcolor{softgreen}{$\downarrow$0.05}) \\

\midrule
~ \textit{w/o} Feature Degradation &  
87.01\%~(\textcolor{softgreen}{$\downarrow$0.11}) & 
91.68\%~(\textcolor{softgreen}{$\downarrow$0.08}) & 
\cellcolor{purple!15}54.32\%~(\textcolor{softgreen}{$\downarrow$0.59}) &  
\cellcolor{purple!15}64.01\%~(\textcolor{softgreen}{$\downarrow$0.90}) \\

~ \textit{w/o} Data Degradation &  
87.11\%~(\textcolor{softgreen}{$\downarrow$0.01}) & 
91.69\%~(\textcolor{softgreen}{$\downarrow$0.07}) & 
\cellcolor{purple!15}54.08\%~(\textcolor{softgreen}{$\downarrow$0.83}) &  
\cellcolor{purple!15}63.97\%~(\textcolor{softgreen}{$\downarrow$0.94}) \\

~ \textit{w/o} Both &  
86.99\%~(\textcolor{softgreen}{$\downarrow$0.13}) & 
91.50\%~(\textcolor{softgreen}{$\downarrow$0.26}) & 
\cellcolor{purple!15}53.95\%~(\textcolor{softgreen}{$\downarrow$0.96}) &  
\cellcolor{purple!15}63.67\%~(\textcolor{softgreen}{$\downarrow$1.24}) \\   
\midrule
\bottomrule
\end{tabular}}}
\vskip -0.1in
\end{table}

\textbf{Controllable Video Generation.}  
Efforts in controllable video generation~\cite{burgert2025gowiththeflow, shi2024motion} have already enabled the synthesis of high-quality videos. VideoComposer~\cite{wang2023videocomposer} leverages diverse conditions to enhance controllability. Following the design of ControlNet~\cite{controlnet}, SparseCtrl~\cite{guo2024sparsectrl} introduces sparse control for video generation, while DAS~\cite{gu2025das} employs 3D point maps for more precise control. Cosmos-Transfer-1~\cite{cosmostransfer1} uses multi-modal control to improve the quality of controllable video generation. However, these methods focus on short clips and do not address the challenges of long-form generation.

\textbf{Long Video Generation.} 
FreeNoise~\cite{qiu2023freenoise} and PYoCo~\cite{ge2023preserve} have explored different noise initialization strategies to enhance the temporal consistency of generated videos. These methods significantly improve overall video consistency. However, fine-grained temporal consistency issues still persist in long video generation. To address this, we propose the use of Unified Noise Initialization in this work.
Additionally, to mitigate the high computational resource requirements, several methods~\cite{cai2025ditctrl,lu2024freelong,tan2025freepca,li2024vstar,zhao2025riflex} have adopted training-free strategies. These include analyzing attention mechanisms or identifying key components in video models to extend the duration of generated videos.
Furthermore, some training-based approaches focus on enhancing video length. StreamingT2V~\cite{StreamingT2V} adopts an autoregressive approach for long video generation by leveraging a short-term memory mechanism to maintain temporal consistency. Diffusion Forcing~\cite{chen2024diffusionforcing, song2025historyguidedvideodiffusion} introduces an autoregressive framework that enables diffusion models to generate long videos through causal learning. This effectively captures temporal dependencies. TTT~\cite{TTT} and LCT~\cite{LCT} explicitly incorporate long-range temporal context to enhance generation quality and coherence across extended sequences.
FramePack~\cite{zhang2025framepack} compresses conditional or previously generated frames into a compact representation as a new condition, emphasizing recent frames through reduced spatial resolution. This improves efficiency and consistency in long video generation.
Although these methods improve long video generation from various perspectives, achieving fine-grained control and maintaining long-term coherence remain open challenges.

\section{Conclusion}
In this work, we investigate the causes of temporal inconsistency and visual degradation in controllable long video generation. To address these issues, we propose \textbf{LongVie}, a multi-modal guidance framework that integrates dense and sparse control signals in an autoregressive manner, supported by a degradation-aware training strategy to enhance visual quality. It also applies global normalization to control signals and uses unified noise initialization to improve temporal consistency. To evaluate controllable long video generation, we curate \textbf{LongVGenBench}, a dataset of 100 high-quality videos, each lasting over one minute and spanning real-world and game-like scenarios. Experiments on LongVGenBench and ablation studies show that LongVie achieves state-of-the-art performance in long video generation. Moreover, downstream video generation tasks demonstrate that LongVie can generate high-quality controllable videos up to one minute long.

\newpage
%%%%%%%%%%%%%%%%%%%%%%%%%%%%%%%%%%%%%%%%%%%%%%%%%%%%%%%%%%%%

\appendix

\section{Overview of Supplementary Material}
In this section, we briefly introduce the contents of the supplementary material.
In Sec.\ref{LongVGenBench}, we provide a detailed description of the \textbf{LongVGenBench} dataset collected for evaluation.
In Sec.\ref{Implementation}, we elaborate on the implementation details of \textbf{LongVie}, including training data, inference adaptation, and model configuration.
In Sec.\ref{Ablation}, we present ablation studies on the structure and number of MM-Blocks, as well as additional explorations of LongVie under varying initial noise and inaccurate control signals.
In Sec.~\ref{more_res}, we show additional qualitative results generated by our model in various styles.

\section{LongVGenBench}
\label{LongVGenBench}
\begin{figure}[h]
\vskip -0.05in
\begin{center}
\includegraphics[width=0.95\columnwidth]{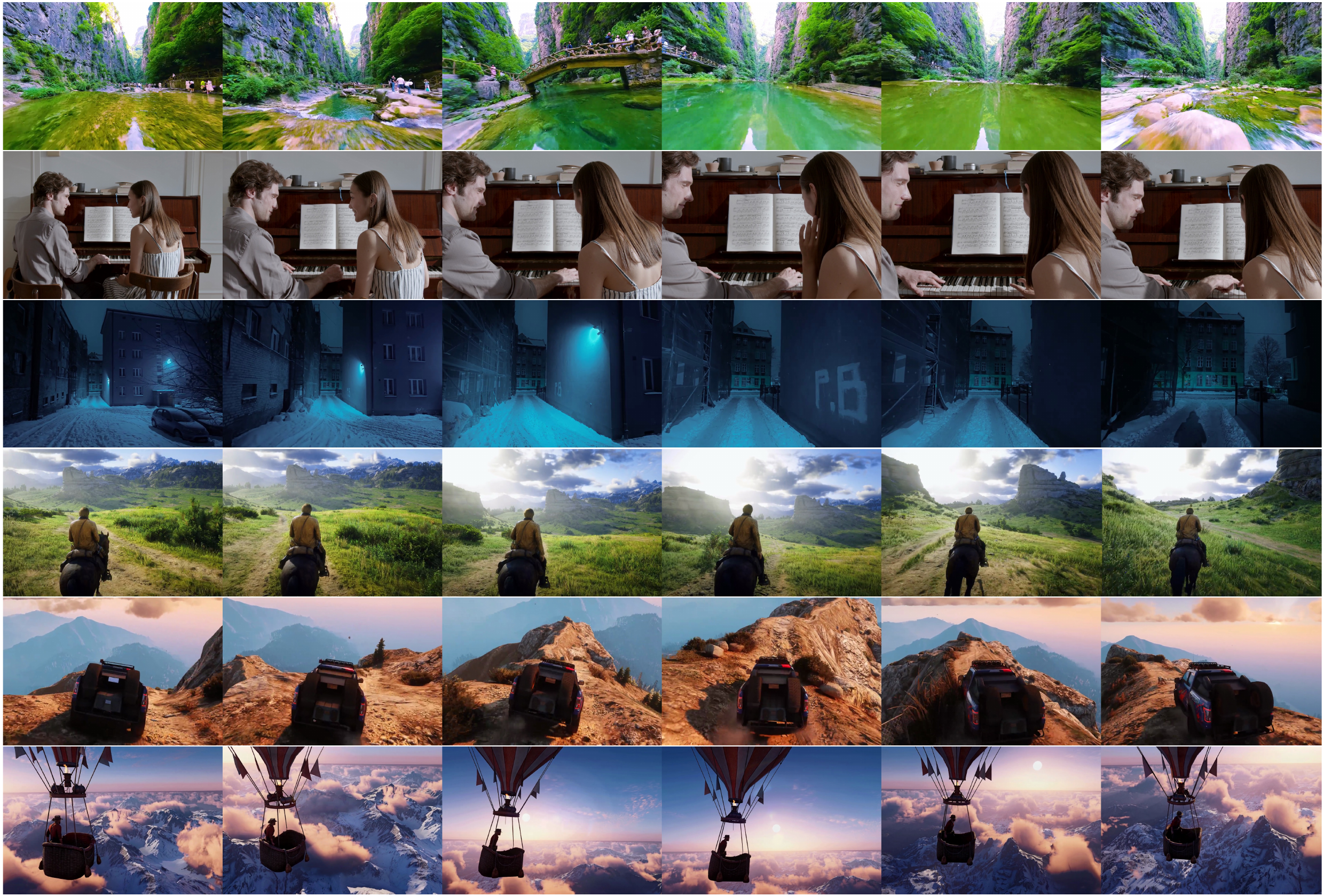}
\vskip -0.05in
\caption{\textbf{Examples from LongVGenBench.} We show several videos from both real-world and synthetic scenarios in LongVGenBench, covering a variety of indoor and outdoor environments to evaluate the controllable long video generation ability of our model.}
\label{fig:longbench}
\end{center}
\vskip -0.1in
\end{figure}

To better introduce the proposed dataset, \textbf{LongVGenBench}, for evaluating controllable long video generation, we present several examples in Fig.~\ref{fig:longbench}. The dataset includes both real-world and synthetic scenes, each lasting at least one minute and having a minimum resolution of 1080p. 
As shown in the figure, the dataset contains various types of camera movements, which pose significant challenges for video generation models. Importantly, LongVGenBench is designed to be model-agnostic and can be used to evaluate any long video generation method, not just LongVie. 
In this paper, we utilize LongVGenBench by splitting each video into 6-second clips with a one-frame overlap. Captions are then extracted from these short clips to prepare the training data. To further illustrate the distribution of the dataset, we provide detailed statistics in Tab~\ref{tab:data_distribution}.

\begin{table}[h]
\small
\centering
\setlength{\tabcolsep}{4mm}{
\caption{Statistics of LongVGenBench video categories.}
\label{tab:data_distribution}
\vskip -0.05in
\begin{tabular}{lccc}
\toprule
 & \textsc{Day} & \textsc{Night} & \textsc{Total} \\
\midrule
\textbf{Real} & 55 & 10 & 65 \\
\quad Indoor & 22 & 0 & 22 \\
\quad Outdoor & 33 & 10 & 43 \\
\textbf{Game} & 25 & 10 & 35 \\
\quad Indoor & 10 & 0 & 10 \\
\quad Outdoor & 15 & 10 & 25 \\
\midrule
\textbf{Total} & 80 & 20 & \textbf{100} \\
\bottomrule
\end{tabular}}
\vskip -0.1in
\end{table}

\section{More Implementation Details}
\label{Implementation}

\paragraph{Training Data.}
As discussed in the \textit{Implementation Details} section of the main paper, our training data consists of three parts: \textit{1) ACID and ACID-Large~\cite{acid}:} These datasets contain thousands of aerial drone videos featuring various coastlines and natural scenes sourced from YouTube. They are released in the same format as RealEstate10K~\cite{RealEstate10K}. \textit{2) Vchitect\_T2V\_DataVerse~\cite{dataverse}:} This dataset comprises 14 million high-quality videos collected from the Internet, each paired with detailed textual captions. \textit{3) MovieNet~\cite{huang2020movienet}:} MovieNet includes 1,100 movies spanning a diverse range of years, countries, and genres.

\paragraph{Data Pre-processing.}
During our experiments, we observed that if the training data contains transitions (e.g., cuts or scene changes), the temporal consistency of the generated videos tends to degrade, often resulting in unintended transitions in the output. To address this, we use the PySceneDetect toolkit~\cite{castellano2024pyscenedetect} to detect transitions in the videos and segment them accordingly. Each resulting segment is then sampled at 8 fps and truncated to 49 frames for training LongVie. From these 49-frame clips, we use Video Depth Anything~\cite{video_depth_anything} to extract depth maps, and compute point tracking using SpatialTracker~\cite{SpatialTracker}, based on the extracted depth and RGB frames. We also generate captions for each clip using Qwen-2.5-VL-7B~\cite{Qwen2.5-VL} to ensure accurate and context-relevant descriptions. In total, we prepare 150,000 video–control signal pairs to train our model.

\begin{figure}[h]
\begin{center}
\includegraphics[width=\columnwidth]{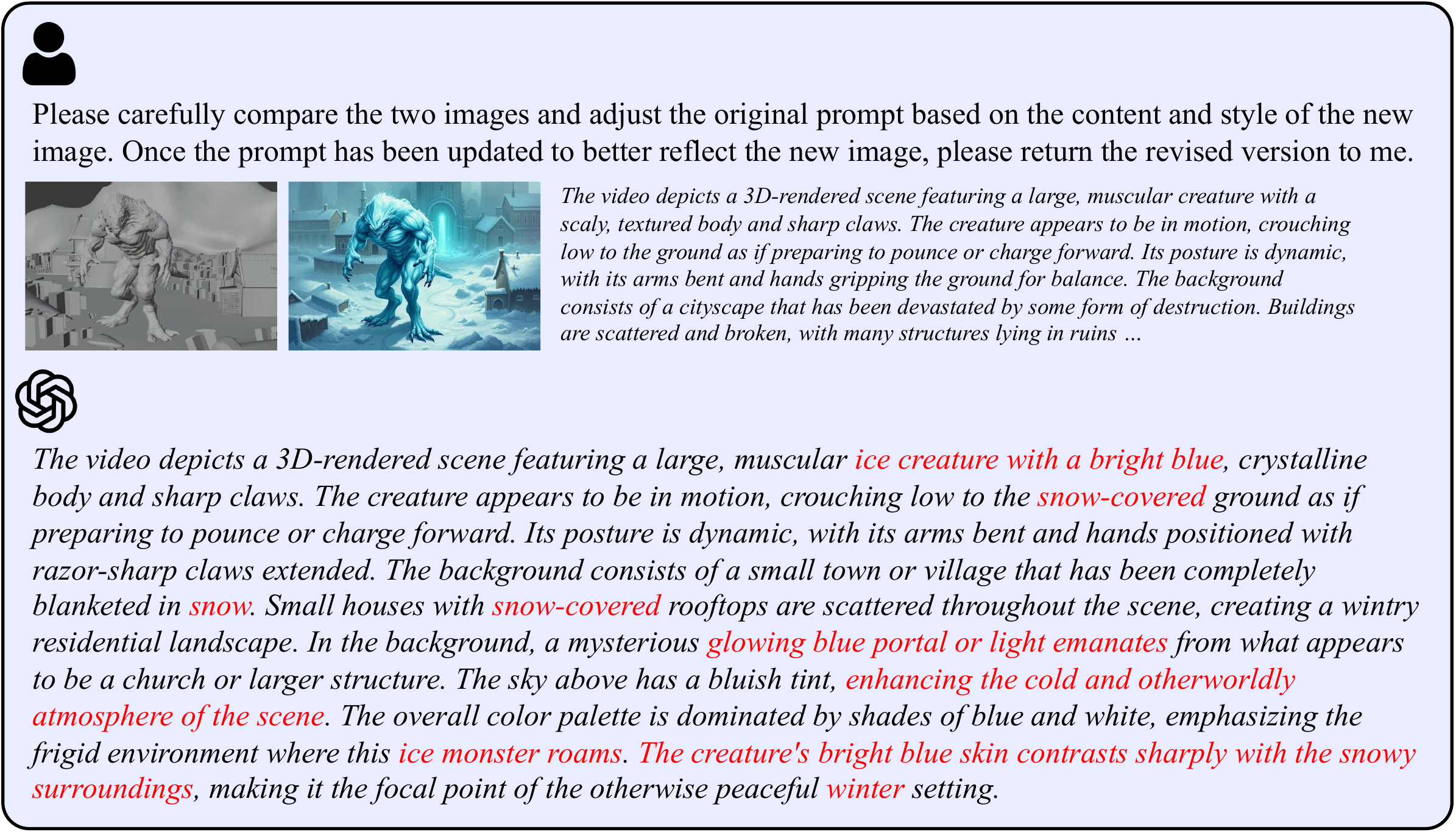}
\vskip -0.05in
\caption{\textbf{Caption Refinement via LLM.} We revise the original captions by comparing the transferred and source images using a LLM. This process ensures that the updated captions accurately reflect the visual content of the transferred video.}
\label{fig:update_caption}
\end{center}
% \vskip -0.1in
\end{figure}

\paragraph{Test Time Adaptation.}
Point tracking becomes less effective during the inference of long videos because it relies on content visible in the first frame. 
As the video progresses, or when the initially tracked content moves entirely out of view, the tracked points tend to fail. In our framework, point tracking serves as a sparse control signal that primarily guides motion rather than appearance. To ensure its effectiveness, we avoid extracting point tracks from the entire 1-minute video directly. Instead, we first extract the depth map for the full 1-minute sequence and apply global normalization. We then divide the video into overlapping 49-frame clips and compute colorful point tracking for each clip based on the globally normalized depth maps.
Regarding captions, in transfer tasks, the original captions of the source videos are often misaligned with the visual content after transfer. To address this, we employ a large language model (LLM) to analyze the differences between the transferred and original frames, and revise the captions to better reflect the updated content, as illustrated in Fig.~\ref{fig:update_caption}.

\paragraph{Model Configuration.}
We provide additional details about our model, LongVie. During training, we set the feature-level degradation probability $\alpha$ to 15\% and the data-level degradation probability $\beta$ to 10\%. One or both degradation methods are randomly selected at each step. We set $n = 5$ for the \textit{Random Scale Fusion} method. The model is trained without any degradation strategies for the first 2000 iterations, and these strategies are introduced in the final 1000 iterations.
For weight initialization, the dense and sparse blocks within each MM Block copy only half of the original weights, and the feature dimension is also reduced to half compared to that in CogVideoX~\cite{yang2024cogvideox}. Specifically, we split the original weights into two interleaved sets based on their index positions (\eg weights at positions 0, 2, 4 and 1, 3, 5), and assign them to the dense and sparse blocks respectively. The half copy significantly reduces the total number of parameters in the model.

\section{Additional Ablation Studies}
\label{Ablation}
To comprehensively evaluate the effectiveness of LongVie, we conduct extensive ablation experiments.

\paragraph{Control Block Architecture.} As discussed in Sec.~\ref{sec2.2}, we compare two designs for integrating control features with the denoising features. The results, summarized in Tab.~\ref{tab:ablation_proj}, indicate that the unified zero-linear approach outperforms the separate design across all metrics.

\begin{table}
\small
\centering
\setlength{\tabcolsep}{3mm}{
\caption{
\textbf{Ablation study on control block architecture.} Comparison between unified and separate zero-linear designs. The unified approach yields better performance across all the metrics.
\label{tab:ablation_proj}
}
\vskip -0.05in
{\begin{tabular}{lcccccc}
\toprule
\multirow{2}{*}{\textsc{Variant}} & \multicolumn{2}{c}{Consistency} & \multicolumn{2}{c}{Quality} & \multirow{2}{*}{\textsc{SSIM}} & \multirow{2}{*}{\textsc{LPIPS}} \\
                                & Subject        & Background        & Aesthetic & Imaging        &        &         \\
\midrule
Separate Zero-Linear  & 87.01\%          & 91.59\%           & 54.80\%          & 64.83\%          & 0.524          & 0.321          \\
\textbf{Unified Zero-Linear}   & \textbf{87.12\%} & \textbf{91.76\%} & \textbf{54.91\%} & \textbf{64.91\%} & \textbf{0.557} & \textbf{0.290} \\
\bottomrule
\end{tabular}}}
% \vskip -0.1in
\end{table}

\paragraph{Number of Control Blocks.} We also investigate the impact of the number of control blocks in LongVie. Specifically, we train a variant using only 12 control blocks and compare it with our default 18-block setting. As shown in Tab.~\ref{tab:ablation_blocks}, increasing the number of blocks leads to consistent improvements across all metrics.

\begin{table}
\small
\centering
\setlength{\tabcolsep}{4.5mm}{
\caption{
\textbf{Ablation study on the number of control blocks.} We compare different numbers of control blocks to evaluate the efficiency and effectiveness of our model.
\label{tab:ablation_blocks}
}
\vskip -0.05in
{\begin{tabular}{lcccccc}
\toprule
\multirow{2}{*}{\textsc{Blocks}} & \multicolumn{2}{c}{Consistency} & \multicolumn{2}{c}{Quality} & \multirow{2}{*}{\textsc{SSIM}} & \multirow{2}{*}{\textsc{LPIPS}} \\
                                & Subject        & Background        & Aesthetic & Imaging        &        &         \\
\midrule
12                              & 85.79\%          & 90.96\%           & 54.52\%          & 64.14\%          & 0.502          & 0.348          \\
18                              & \textbf{87.12\%} & \textbf{91.76\%} & \textbf{54.91\%} & \textbf{64.91\%} & \textbf{0.557} & \textbf{0.290} \\
\bottomrule
\end{tabular}}}
\vskip -0.1in
\end{table}

\paragraph{Exploration of Initial Noise.} 
To investigate the impact of initial noise on the temporal consistency of generated videos, we evaluate our model under controlled perturbations of the initialization noise. Specifically, we add Gaussian noise sampled from $\mathcal{N}(0,1)$ and scaled by a factor $\alpha$ to the Global Initialization Noise. Temporal consistency is assessed using our proposed LongVGenBench dataset, and we report four key metrics: \textit{Subject Consistency}, \textit{Background Consistency}, \textit{Temporal Style}, and \textit{Overall Consistency}, as shown in Tab.~\ref{tab:noise_comparison_table}. The results indicate that smaller variations in the initial noise lead to improved temporal consistency in the generated videos.

\begin{table}
\centering
\small
\setlength{\tabcolsep}{3mm}{
\caption{Comparison of different initial noise strategies.}
\label{tab:noise_comparison_table}
\vskip -0.08in
\begin{tabular}{lcccc}
\toprule
\multirow{2}{*}{\textsc{Method}} & Subject & Background & Overall & Temporal \\
                                 & Consistency & Consistency & Consistency  & Style \\
\midrule
Random Noise & 86.39\% & 91.41\% & 21.17\% & 21.17\% \\
Global Initial Noise with $\alpha=0.5$ & 86.51\% & 91.48\% & 21.25\% & 21.35\% \\
Global Initial Noise with $\alpha=1$ & 86.83\% & 91.62\% & 21.46\% & 21.56\% \\
Global Initial Noise with $\alpha=0.05$ & 86.96\% & 91.66\% & 21.64\% & 21.74\% \\
Global Initial Noise & \textbf{87.12\%} & \textbf{91.76\%} & \textbf{21.76\%} & \textbf{21.82\%} \\
\bottomrule
\end{tabular}}
\vskip -0.05in
\end{table}

\paragraph{Robustness to Inaccurate Control Signals.} 
To further assess the robustness of our method, we evaluate its performance under degraded control conditions, including depth maps blurred with a $5 \times 5$ kernel and point maps with 20\% of keypoints randomly masked. As shown in Tab.~\ref{tab:control_modalities}, our method achieves performance comparable to the clean-control setting, demonstrating strong resilience to control signal degradation with only minimal performance drop.

\begin{table}
\centering
\small
\setlength{\tabcolsep}{1mm}{
\caption{Evaluation of different control modalities across multiple video generation metrics.}
\vskip -0.08in
\label{tab:control_modalities}
\begin{tabular}{lccccccccc}
\toprule
Method & S.C. & B.C. & O.C. & D.D. & T.F. & A.Q. & I.Q. & SSIM & LPIPS \\
\midrule
Masked Keypoint Map  & 86.74\% & 91.28\% & 21.54\% & 46.18\% & 97.71\% & 54.63\% & 64.52\% & 0.531 & 0.312 \\
Blurred Depth Map    & 86.91\% & 91.42\% & 21.68\% & 46.42\% & 97.94\% & 54.75\% & 64.75\% & 0.548 & 0.297 \\
LongVie              & \textbf{87.12\%} & \textbf{91.76\%} & \textbf{21.82\%} & \textbf{46.59\%} & \textbf{98.43\%} & \textbf{55.31\%} & \textbf{64.91\%} & \textbf{0.557} & \textbf{0.290} \\
\bottomrule
\end{tabular}}
\end{table}

\section{More Qualitative Results}
\label{more_res}
\begin{figure}[h]
% \vskip -0.1in
\begin{center}
\includegraphics[width=\columnwidth]{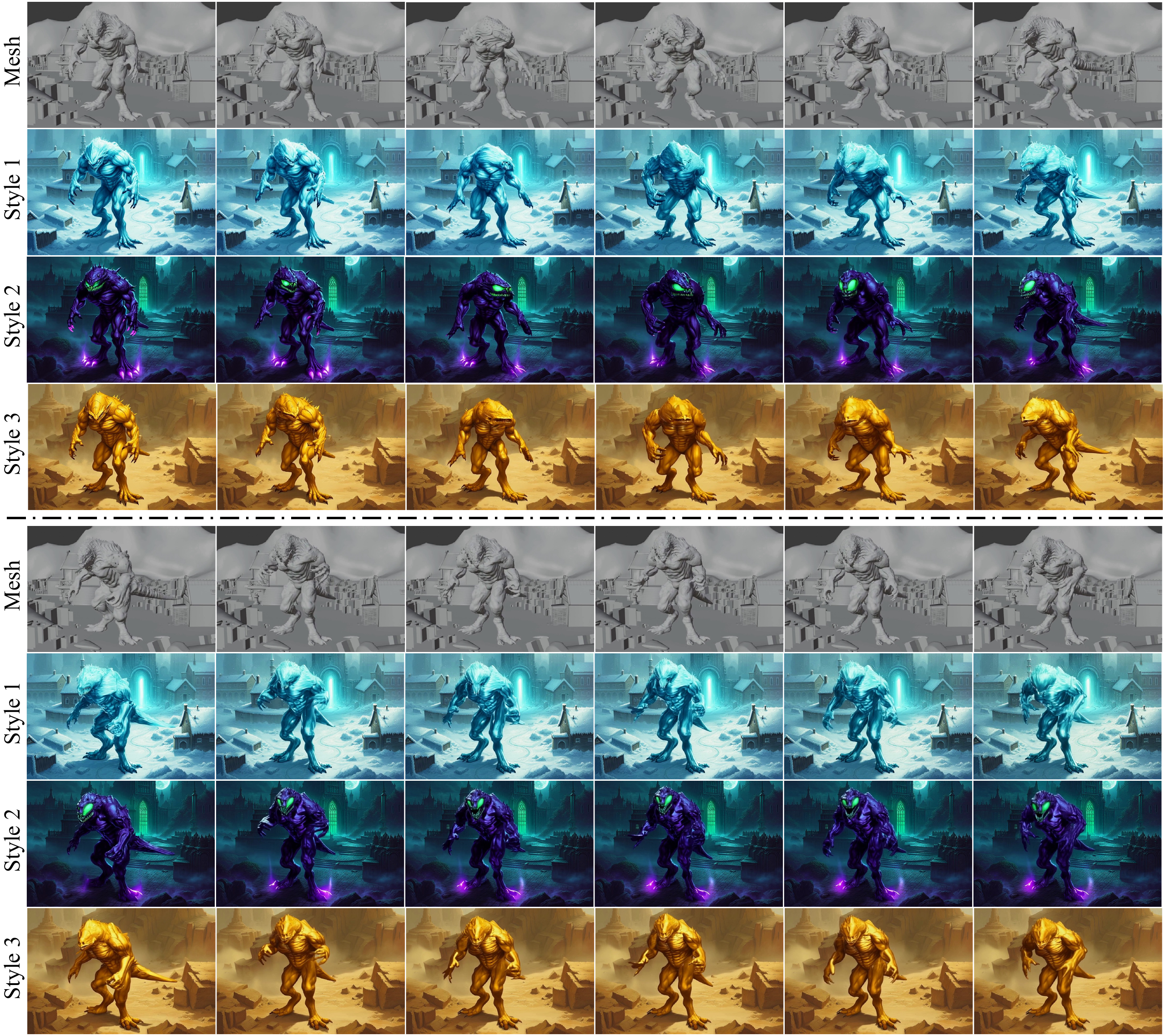}
% \vskip -0.1in
\caption{\textbf{More results of Mesh-to-Video.} We input a monster located in a village scene, animate the mesh, and convert it into a video, where our model supports rendering in various styles.}
\label{fig:more_mesh2video}
\end{center}
% \vskip -0.2in
\end{figure}

\begin{figure}[h]
\begin{center}
\includegraphics[width=\columnwidth]{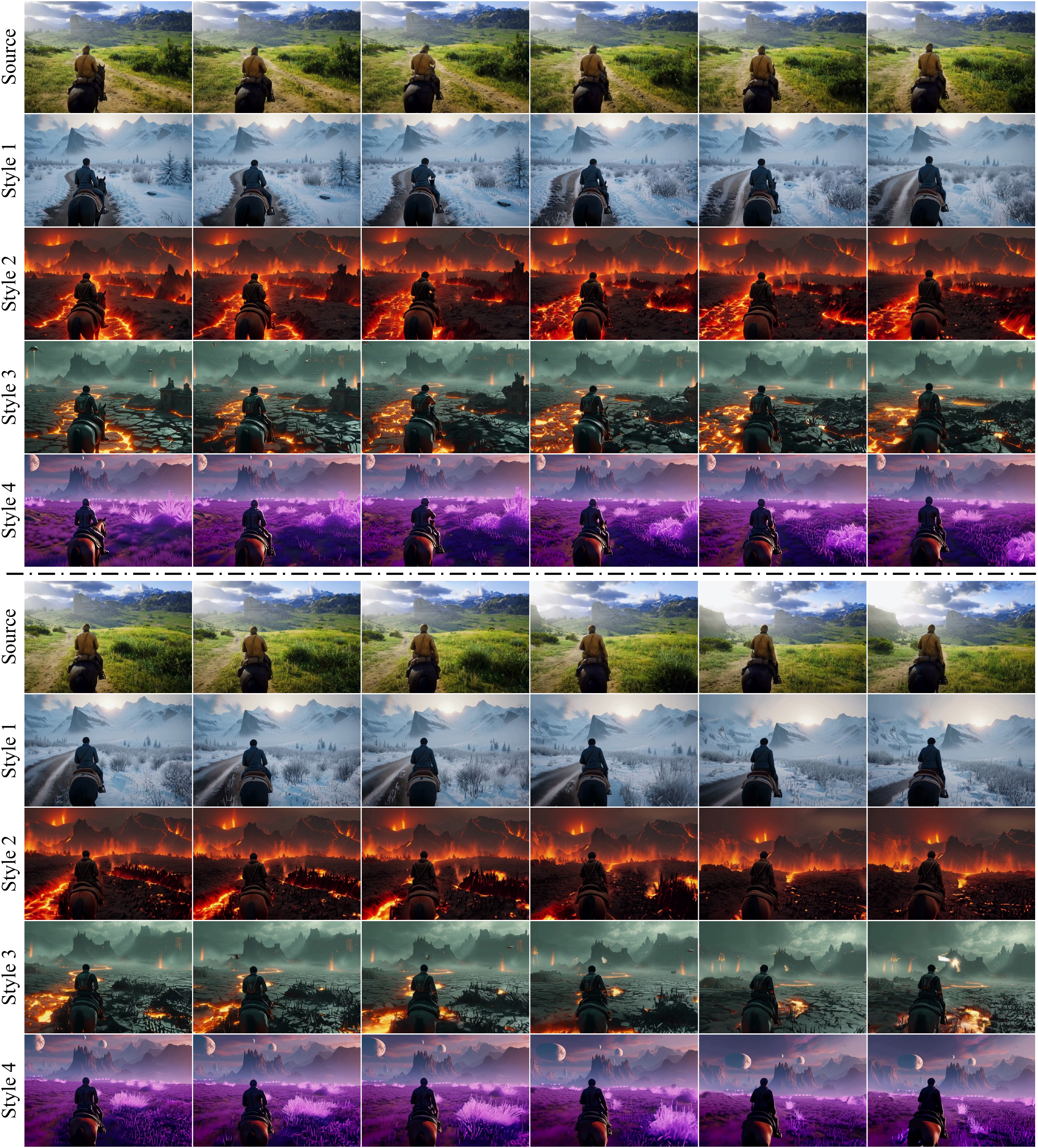}
\vskip -0.05in
\caption{\textbf{More results of Motion \& Scene Transfer.} A man riding a horse in various scenes.}
\label{fig:suppl_transfer}
\end{center}
% \vskip -0.2in
\end{figure}

\begin{figure}
\begin{center}
\includegraphics[width=\columnwidth]{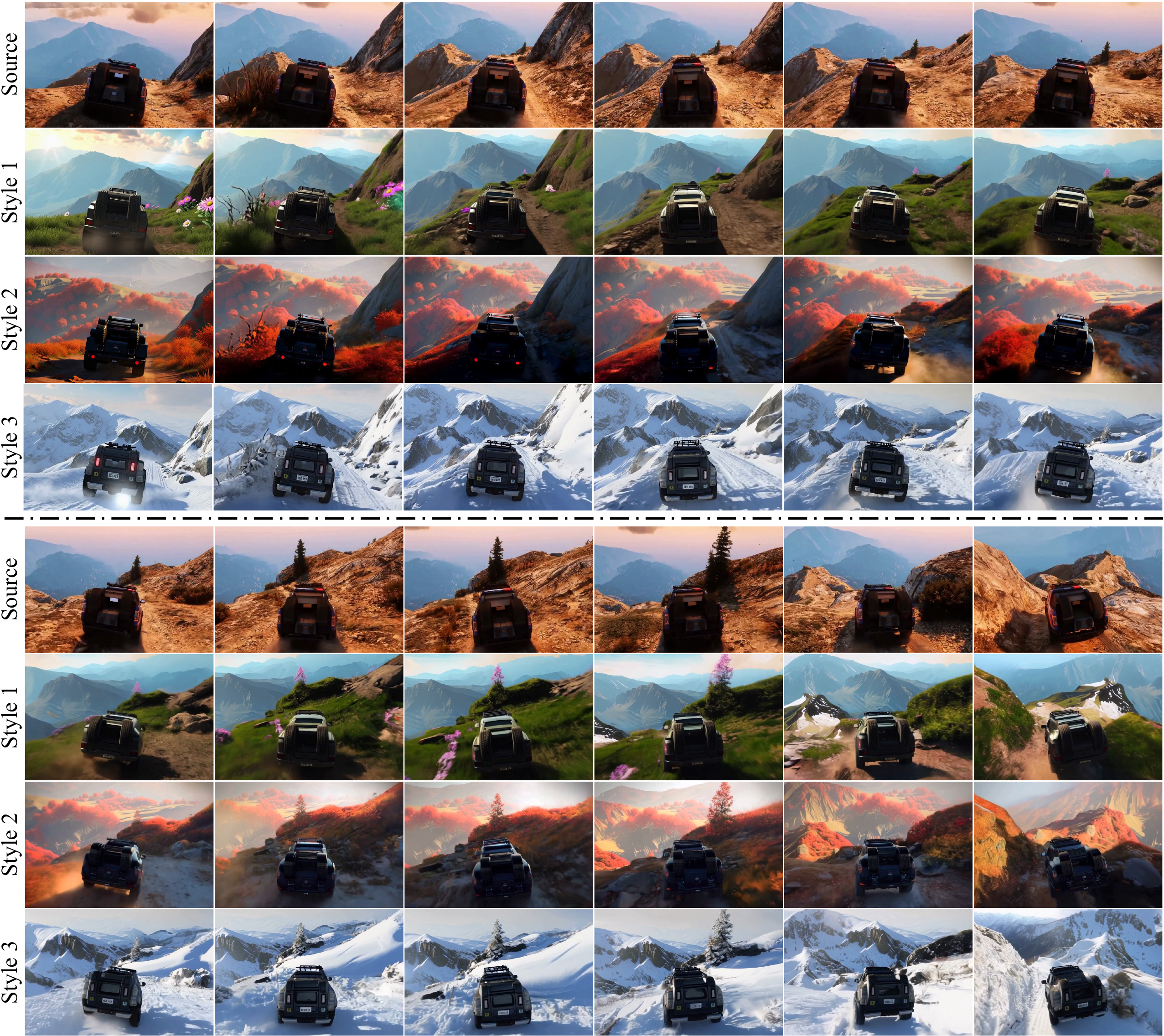}
\vskip -0.05in
\caption{\textbf{More results of Motion \& Scene Transfer.} We transfer a car driving through a mountain valley across various seasons.}
\label{fig:suppl_transfer2}
\end{center}
\vskip -0.1in
\end{figure}

To better demonstrate the robustness of our model, we present additional experimental results on various long video generation tasks. The results are shown in Fig.~\ref{fig:more_mesh2video}, Fig.~\ref{fig:suppl_transfer} and Fig.~\ref{fig:suppl_transfer2}.

\clearpage

\clearpage

\section{Social Impacts}

LongVie facilitates more customized video generation, with potential applications across various fields such as short-form content creation and the film industry. By enhancing controllability and consistency in generated videos, it can significantly streamline video production processes and expand creative possibilities, though it is also crucial to consider and address potential risks, including misinformation or misuse.

\section{Limitations and Future Work}

Although LongVie is capable of generating high-quality, controllable 1-minute videos, the inference process remains relatively time-consuming, requiring approximately 45 minutes per video at 480$\times$720 resolution. Reducing inference latency for long video generation remains a non-trivial challenge and is an important direction for future research.
Furthermore, the current output resolution, while sufficient for benchmarking, still falls short of cinematic standards. In future work, we aim to explore higher-resolution generation frameworks to produce visually richer and more realistic long videos.

\clearpage

{\small
\bibliographystyle{plain}
\bibliography{egbib}
}

\end{document}